\documentclass[letterpaper]{article} %
\usepackage{aaai2026}  %
\usepackage{times}  %
\usepackage{helvet}  %
\usepackage{courier}  %
\usepackage[hyphens]{url}  %
\usepackage{graphicx} %
\usepackage{natbib}  %
\usepackage{caption} %
\usepackage{algorithm}
\usepackage{algorithmic}
\usepackage{booktabs}
\usepackage{newfloat}
\usepackage{listings}
\DeclareCaptionStyle{ruled}{labelfont=normalfont,labelsep=colon,strut=off} %
\floatstyle{ruled}
\newfloat{listing}{tb}{lst}{}
\floatname{listing}{Listing}
\title{VoxRole: A Comprehensive Benchmark for Evaluating Speech-Based Role-Playing Agents}
\author{
    Weihao Wu\textsuperscript{\rm 1}\equalcontrib, Liang Cao\textsuperscript{\rm 1}\equalcontrib, Xinyu Wu\textsuperscript{\rm 1}\equalcontrib, Zhiwei Lin\textsuperscript{\rm 1}, Rui Niu\textsuperscript{\rm 1}, \\Jingbei Li\textsuperscript{\rm 2}, Zhiyong Wu\textsuperscript{\rm 1}\thanks{Corresponding author}\\
}
\affiliations{
    \textsuperscript{\rm 1}Shenzhen International Graduation School, Tsinghua University\\
    \textsuperscript{\rm 2}StepFun\\
    \{wuwh23, cao-l24, xy-wu24\}@mails.tsinghua.edu.cn, zywu@sz.tsinghua.edu.cn
}

\usepackage{bibentry}

\begin{document}

\maketitle

\begin{abstract}
Recent significant advancements in Large Language Models (LLMs) have greatly propelled the development of Role-Playing Conversational Agents (RPCAs). These systems aim to create immersive user experiences through consistent persona adoption.
However, current RPCA research faces dual limitations. First, existing work predominantly focuses on the textual modality, entirely overlooking critical paralinguistic features including intonation, prosody, and rhythm in speech, which are essential for conveying character emotions and shaping vivid identities. Second, the speech-based role-playing domain suffers from a long-standing lack of standardized evaluation benchmarks. Most current spoken dialogue datasets target only fundamental capability assessments, featuring thinly sketched or ill-defined character profiles. Consequently, they fail to effectively quantify model performance on core competencies like long-term persona consistency.
To address this critical gap, we introduce VoxRole, the first comprehensive benchmark specifically designed for the evaluation of speech-based RPCAs. The benchmark comprises 13335 multi-turn dialogues, totaling 65.6 hours of speech from 1228 unique characters across 261 movies. To construct this resource, we propose a novel two-stage automated pipeline that first aligns movie audio with scripts and subsequently employs an LLM to systematically build multi-dimensional profiles for each character. Leveraging VoxRole, we conduct a multi-dimensional evaluation of contemporary spoken dialogue models, revealing crucial insights into their respective strengths and limitations in maintaining persona consistency.
\end{abstract}

\section{Introduction}
The advent of Large Language Models (LLMs) has catalyzed a paradigm shift in conversational Artificial Intelligence, enabling the development of systems that exhibit unprecedented fluency and contextual understanding in human-machine interaction \cite{achiam2023gpt, kosinski2023theory, wang2024survey}. A particularly compelling frontier within this domain is the development of Role-Play Conversational Agents (RPCAs), which have garnered significant academic and commercial attention \cite{shanahan2023role}. Unlike standard task-oriented or open-domain chatbots, RPCAs are distinguished by their core capability to consistently adopt and maintain a predefined persona or character. This ability to embody a specific identity—be it a historical figure, a fictional character, or a therapeutic guide—is paramount for cultivating immersive, engaging, and authentic conversational experiences. Consequently, these agents hold immense potential across a diverse array of applications, including interactive entertainment, personalized education, and mental health support.
\begin{figure}[htbp] %
\centering
\includegraphics[width=0.98\columnwidth]{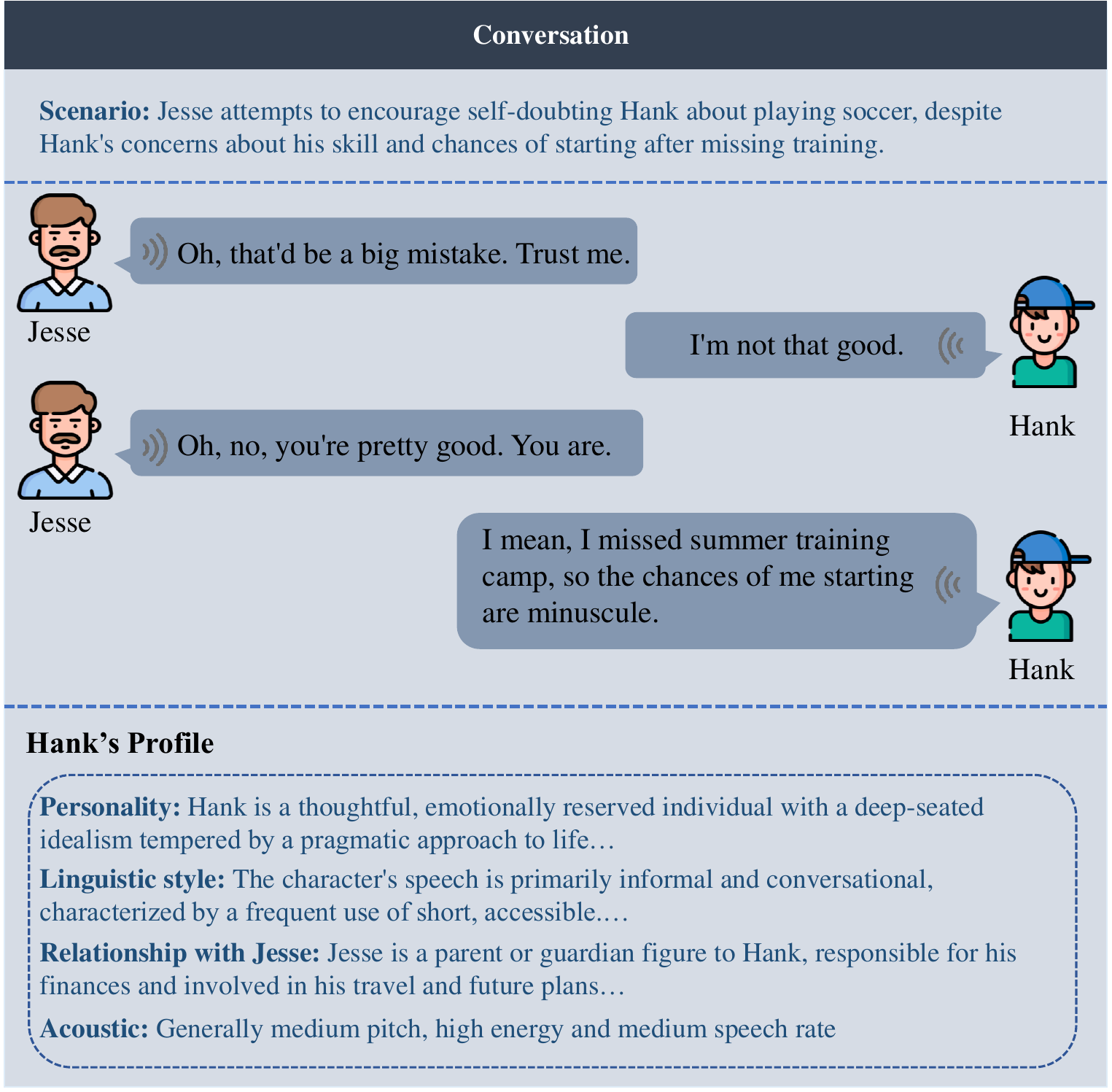} 
\caption{An example of the VoxRole benchmark}
\label{fig:single_column_example}
\end{figure}
 
However, the prevailing trajectory of RPCA research has been overwhelmingly concentrated on the textual modality \cite{wang2023rolellm, liu2024roleagent, tu2024charactereval, wang2025coser}, thereby overlooking the critical dimension of speech. This focus on text-based interaction, while foundational, imposes a significant limitation on the achievable level of realism and emotional depth. The textual medium is inherently constrained, stripping communication of vital paralinguistic and suprasegmental cues such as prosody, intonation, pitch, rhythm, and vocal timbre. These auditory signals are fundamental to human communication, serving as the primary channels for conveying emotional subtleties, indicating sarcasm, expressing confidence, and ultimately shaping a character's unique and memorable identity. Speech, in stark contrast, naturally excels at translating abstract personality traits and transient emotional states into these perceptible auditory signatures. Therefore, the integration of speech is not merely an incremental enhancement but a fundamental prerequisite for RPCAs to transcend their current limitations and achieve true verisimilitude.

Despite this clear imperative, progress toward robust, speech-based RPCAs is significantly impeded by a critical bottleneck: the scarcity of high-quality, standardized benchmarks for evaluation. Existing spoken dialogue benchmarks \cite{cheng2025voxdialogue, chen2024voicebench, yan2025uro, lin2025full}, while instrumental in advancing and assessing foundational competencies such as automatic speech recognition, language understanding, and logical reasoning, are fundamentally inadequate for evaluating the nuanced art of role-playing. This deficiency stems primarily from their lack of spoken dialogue data enriched with well-defined, multi-dimensional character profiles that detail personality, background, and relationships. At the root of this systemic issue is the field's historical reliance on manual annotation for dataset creation—a methodology that is not only prohibitively expensive and labor-intensive but also inherently unscalable and prone to subjective inconsistencies. The resulting absence of a rigorous and systematic evaluation framework for speech-based role-playing creates a vicious cycle: it prevents a comprehensive understanding of current model capabilities, obscures meaningful progress, and ultimately obstructs the targeted development of more contextually and socially aware dialogue agents.

To address this critical gap, we introduce VoxRole, the first comprehensive benchmark specifically designed for speech-based RPCAs. The construction of VoxRole is enabled by a novel, two-stage automated pipeline designed for the efficient extraction of character-rich spoken dialogues from movies. Initially, the pipeline processes and aligns movie audio tracks with their corresponding scripts to generate a corpus of speaker-annotated dialogues. Subsequently, an LLM is leveraged to analyze this corpus, systematically constructing a multi-dimensional profile for each character, encompassing personality traits, linguistic style, interpersonal relationships, and acoustic characteristics. This process yielded the VoxRole benchmark, which comprises 13335 two-speaker, multi-turn dialogues (approx. 65.6 hours of speech) and features 1228 unique characters across 261 movies. Leveraging this new resource, we conduct a multi-dimensional evaluation of contemporary large-scale spoken dialogue models, revealing several insightful findings regarding their capabilities and limitations.

To summarize, the main contributions of this paper are:

\begin{itemize}
    \item We establish VoxRole, a comprehensive benchmark for speech-based RPCAs, which introduces a systematic evaluation framework.
    \item We have developed a two-stage pipeline to automatically extract spoken dialogues with speaker annotations and persona information without manual labeling. 
    \item We perform a thorough evaluation of existing large-scale spoken dialogue models using VoxRole. The analysis yields valuable insights into their current capabilities and limitations in the context of role-playing, as well as provides a strong baseline for future research.
\end{itemize}

\begin{table}[t]

  \centering
  \begin{tabular}{lccc}
    \toprule
    \textbf{For each movie} &\textbf{Min} &\textbf{Avg} &\textbf{Max} \\
    \midrule
    Number of characters & 1 & 4.7 & 11 \\
    Number of conversations & 7 & 51.1 & 177 \\
    Duration & 5.1 min & 15.1 min & 62.2 min \\
    \toprule
    \textbf{For each conversation} &\textbf{Min} &\textbf{Avg} &\textbf{Max} \\
    \midrule
    Number of sentences & 3 & 4.44 & 37 \\
    Duration & 3.5 s & 17.7 s & 6.1 min \\
    \toprule
    \textbf{For each character} &\textbf{Min} &\textbf{Avg} &\textbf{Max} \\
    \midrule
    Number of sentences & 10 & 40.0 & 344 \\
    Duration & 25.2 s & 157.5 s & 24.5 min \\
    \bottomrule
  \end{tabular}
  \caption{Statistics of the VoxRole benchmark}
  \label{tab:dataset}
\end{table}

\section{Related Work}
The integration of RPCAs and spoken dialogue systems is advancing next-generation immersive human-computer interaction. While text-based RPCAs have demonstrated substantial progress in persona consistency, their transition to speech-enabled agents confronts two critical challenges: first, the absence of paralinguistic features such as prosody and emotional timbre, which are essential for conveying character identity; and second, the lack of standardized evaluation benchmarks capable of quantifying nuanced spoken role-playing performance. Current speech models predominantly focus on multi-task comprehension and real-time responsiveness, resulting in severely limited research on character profile-driven speech interactions. This gap not only constrains applications like interactive storytelling but also impedes technological advancement in highly expressive speech-based RPCAs.
\begin{figure*}[t]
\centering
\includegraphics[width=1\textwidth]{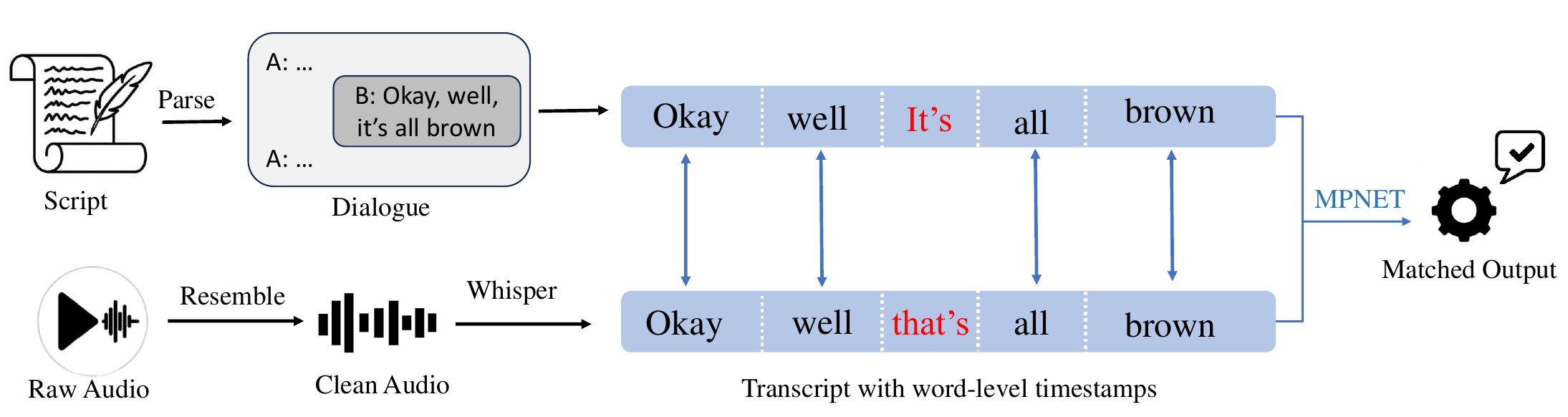} %
\caption{The automatic spoken dialogue extraction pipeline}
\label{fig:sd_pipeline}
\end{figure*}
\subsection{Role-Play Conversational Agents}

The development of role-playing research clearly demonstrates phased technological evolution.
Early work primarily focused on establishing evaluation systems, where  systematically assessed LLMs' performance in factual consistency and motivation recognition through the CROSS dataset, laying the foundation for subsequent studies. 
The field then entered a capability enhancement phase with the RoleLLM framework \cite{wang2023rolellm}, which significantly improved role-playing abilities through instruction tuning and role-specific knowledge injection. 
As the technology matured, \cite{lu2025rolemrc} upgraded evaluation methods by developing the RoleMRC benchmark for fine-grained behavioral analysis. Notably, research focus naturally shifted toward optimizing role depth and attractiveness. 
CharacterEval \cite{tu2024charactereval} introduced human-aligned reward models to quantify "character attractiveness", while CoSER \cite{wang2025coser} enriched role expression dimensions through their "given-circumstance acting" approach for simulating characters' inner states. 
In recent years, the field has expanded toward scalability and multimodal directions: RoleAgent \cite{liu2024roleagent} enabled automatic agent generation, while MM-Role \cite{dai2024mmrole} and  OmniCharacter \cite{zhang2025omnicharacter} explored new paradigms in multimodal interaction and voice dialogue respectively, collectively advancing role-playing technology toward more immersive experiences.

\subsection{Spoken Dialogue System}

The rapid evolution of voice interaction technologies has transitioned from fragmented task-specific solutions toward integrated intelligent ecosystems, where breakthroughs in foundational architectures, interaction paradigms, and multimodal cognition demonstrate mutually reinforcing progress. At the architectural frontier, Qwen2.5-Omni \cite{xu2025qwen2} pioneers end-to-end omnimodal processing through its Thinker-Talker decoupling mechanism and TMRoPE cross-modal alignment, enabling simultaneous text/audio/visual streaming with sub-second latency.

Building upon these infrastructural advances, real-time interaction fidelity has achieved quantum leaps: Baichuan-Audio \cite{li2025baichuan} ’s multi-codebook discretization establishes a new standard for low-latency bilingual dialogue with emotive coherence, a direction further refined by Moshi \cite{defossez2024moshi} ’s full-duplex interruptibility that mirrors human conversational dynamics. Parallel innovations in expressive control emerge through GLM-4-Voice \cite{zeng2024glm} and Step-Audio \cite{huang2025step} ’s prosodic manipulation capabilities, though Qwen2.5-Omni’s streaming DiT decoder transcends temporal constraints by delivering millisecond responsiveness without sacrificing naturalness. Crucially, this technological convergence finds ultimate expression in multimodal cognition systems. GPT-4o \cite{hurst2024gpt} and Qwen2.5-Omni now enable contextual audiovisual reasoning (e.g., vocalizing object relationships derived from live video), while demonstrating that accessibility need not compromise sophistication, MiniCPM-o \cite{team2025minicpm} replicates such capabilities on resource-limited devices – completing an innovation cycle from infrastructure to democratization.

In contrast, only a few studies have explored models’ capabilities in multi-turn role-playing dialogue conditioned on explicit background settings and character profiles, which remains underexplored and holds significant potential for applications in personalized conversational agents and interactive storytelling.

\section{Method}

This chapter delineates the comprehensive methodology developed for the construction of the VoxRole Benchmark and its corresponding evaluation framework. We begin by detailing a fully automated pipeline designed to address the critical challenge of sourcing high-quality, speaker-aligned spoken dialogues from cinematic media. Subsequently, we introduce a novel, LLM-driven approach for generating the rich, multi-faceted character personas that are central to this benchmark. Finally, we establish a robust and multi-dimensional evaluation framework, specifically tailored to assess the nuanced capabilities of conversational agents in role-playing scenarios.

\subsection{Spoken Dialogue Extraction}

To address the scarcity of high-quality spoken dialogues with character annotations, we develop a fully automated pipeline for extracting conversational data from films without manual intervention. We choose movies as the data source because they naturally contain rich character portrayals, diverse emotional expressions and natural dialogue flows. The pipeline comprises three stages: (1) Data Collection and Preparation, (2) Word-level Transcript-Script Alignment, and (3) Semantic Validation and Dialogue Curation. Figure \ref{fig:sd_pipeline} illustrates the complete workflow.

\subsubsection{Data Collection and Preparation}
\begin{figure*}[t]
\centering
\includegraphics[width=0.95\textwidth]{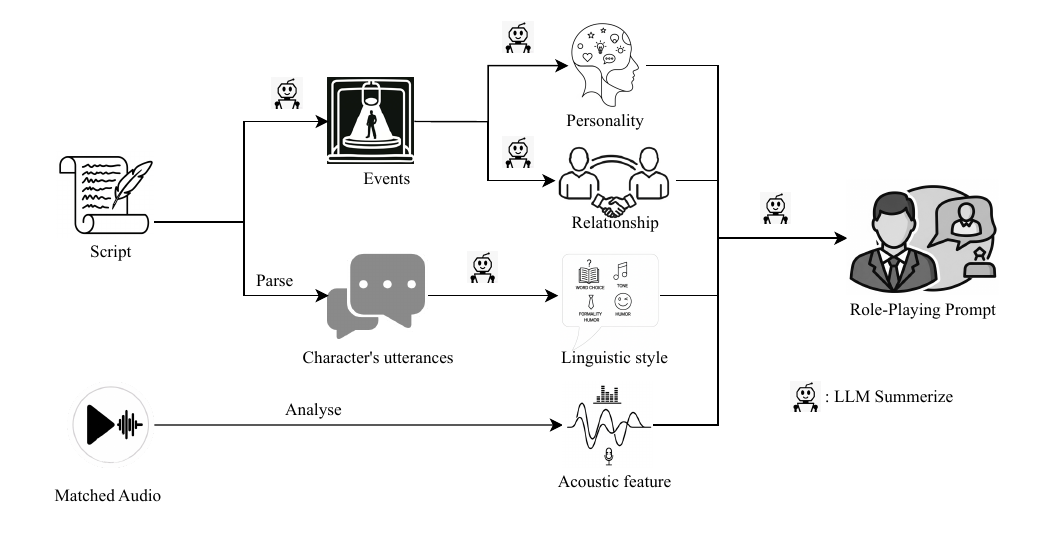} %

\caption{The automatic persona extraction pipeline}

\label{fig:persona_pipeline}
\end{figure*}
Movie scripts were sourced from the Movie Script Database\footnote{https://github.com/Aveek-Saha/Movie-Script-Database}, containing primarily narrative descriptions and speaker-labeled dialogues. Since the original scripts are usually unstructured text, we use regular expressions to parse the scripts into dialogue sequences, each consisting of an utterance and its speaker identifier. It is worth noting that narration is considered as a distinct category of speaker. Then, we collected the corresponding movie video files according to the script metadata (movie title and release year). In order to facilitate the subsequent speech processing, we used the FFmpeg tool to extract the audio tracks of all video files into wav audio files with a sampling rate of 44100Hz.

\subsubsection{Word-Level Audio-Script Alignment}

Word-level audio-acript alignment aims to align the script dialogue with its corresponding segments in the audio stream. We first employ the audio denoising model Resemble \footnote{https://huggingface.co/spaces/ResembleAI/resemble-enhance} to remove noise from movie audio. Audio transcription is then performed using Whisper-large-v3\footnote{https://huggingface.co/openai/whisper-large-v3}, followed by forced alignment with Wav2Vec2.0 to obtain word-level timestamps. To establish a fine-grained correspondence between the transcript and the scripted dialogue, we developed a dynamic matching algorithm, where the original script and the audio transcript are first modeled as word sequences. Subsequently, a word-level alignment is generated by calculating the minimum edit distance between them. Based on this detailed mapping, the precise segment of words in the transcript corresponding to each sentence of the original script can thus be determined.

\subsubsection{Semantic Validation and Dialogue Curation}

A common phenomenon in movies is the divergence between the script and the performed dialogue. Actors frequently paraphrase or improvise, resulting in utterances that are semantically aligned with the script but lexically different. Accordingly, for each candidate sentence pair, we compute its semantic similarity with MPNet \footnote{https://huggingface.co/sentence-transformers/paraphrase-multilingual-mpnet-base-v2}. A pair is confirmed as a successful match if its similarity score exceeds a predefined threshold of 0.8. Subsequently, the set of successful matches is filtered to extract consecutive, two-speaker conversational segments comprising at least three turns. After completing all the matching, we kept the movies with matching audio greater than 5 minutes, and also kept all speakers with more than 10 matched dialogues as characters.

\subsubsection{Benchmark Statics}

The statistical profile of the VoxRole dataset provides insights into the performance of our underlying audio-script alignment framework. The distribution of matched movie durations, which ranges from 5.1 to 62.2 minutes with an average of 15.1 minutes, is particularly revealing. The lower bound of 5.1 minutes typically corresponds to films with significant discrepancies between the script and the final cut, which naturally limits the coverage of our alignment algorithm. Conversely, the upper bound of 62.2 minutes demonstrates the framework's robustness and efficacy in handling films with high content fidelity and minimal divergence. For a comprehensive statistical breakdown, including distributions of characters, conversations, and utterances, please refer to Table \ref{tab:dataset}.

\subsection{Persona Distillation}
\begin{table*}[htb]

  \centering
   \begin{tabular}{lccccc}
    \toprule
    
    \textbf{Models} & \textbf{Size} & \textbf{Rouge-L} & \textbf{Meteor} & \textbf{BertScore F1} & \textbf{UTMOS} \\
    \midrule
    Qwen2.5-Omni \cite{xu2025qwen2}  & $$7B$$ & $10.96$ & $14.60$ &$83.42$& $3.57$ \\
    GLM-4-Voice \cite{zeng2024glm} & $$9B$$ &$9.05$ &$13.55$ &$82.81$& $3.35$ \\
    Baichuan-Audio \cite{li2025baichuan} & $$7B$$ &$9.25$ &$7.75$ &$83.63$& $2.82$ \\
    MiniCPM \cite{team2025minicpm} & $$8B$$ &$10.16$ &$8.64$ &$81.27$& $2.08$ \\
    
    Step-Audio \cite{huang2025step} & $$132B$$ & $12.33$ & $11.44$ &$\mathbf{84.16}$& $2.42$ \\
    GPT-4o & -- & $\mathbf{12.91}$ & $\mathbf{18.30}$ &$83.69$& $\mathbf{3.66}$ \\
    Gemini-2.5-flash & -- & $12.07$ & $14.23$ &$83.65$& $3.39$ \\
    \bottomrule
  \end{tabular}
  \caption{
  Metric-based results on VoxRole.
  }
  \label{tab:results}
\end{table*}

In our work, we divide human persona into four aspects: personality, linguistic style, interpersonal relationships, and acoustic characteristics. The rationale for this division is that it systematically constructs a complete persona through a layered approach, progressing from the internal core to its external manifestation. Personality serves as the psychological core, driving the character's motivations and fundamental behaviors \cite{john1999big}. Interpersonal relationships place the character within a social context, defining their identity and shaping their behavioral patterns \cite{goffman2023presentation}. Linguistic style then acts as the primary symbolic expression of this inner self and social standing, translating them into specific linguistic choices \cite{labov1973sociolinguistic}. Finally, acoustic characteristics provide the physical medium through which all of the above is conveyed, giving the persona its tangible, audible presence and making it multi-dimensional and believable. The total pipeline is shown in \ref{fig:persona_pipeline}

\subsubsection{Acoustic feature acquisition}

Initially, we extracted key acoustic features from the dialogue audio corresponding to each character. Specifically, for each character, the mean pitch, energy, and speech rate were computed across all of their respective utterances. Subsequently, the characters were classified based on the distribution of these three features. For each feature, characters were ranked, with the top 20\% categorized as `High,' the bottom 20\% as `Low,' and the middle 60\% as `Medium'.

\subsubsection{LLM-based persona extraction}
We propose an LLM-based methodology to automatically extract character attributes from raw screenplays without relying on manual annotation. A key challenge is that while screenplays contain a wealth of character-related information, the data useful for defining a character's persona is sparse, rendering direct extraction by a Large Language Model (LLM) inefficient. To address this, inspired by prior work \cite{liu2024roleagent}, we designed a two-stage distillation framework for personality and relationship extraction. Initially, the screenplay is segmented by scene, with overly long scenes further partitioned into smaller chunks. An LLM then summarizes the core events within each segment. Subsequently, for each character, these event summaries are aggregated to infer their personality traits. If two characters co-participate in a sufficient number of events, their shared event history is used to deduce their relationship. This event-summarization process inherently discards nuances of dialogue. Therefore, to capture linguistic style, we adopt a separate approach: we parse and collect all dialogue lines for each character directly from the script and employ an LLM to generate a summary of their unique linguistic style.

After obtaining the complete character profile, we synthesized its four key dimensions nto a cohesive, second-person narrative. This narrative then served as the direct role-playing prompt for the LLM.

\subsubsection{Quality Validation}

To validate the effectiveness of our character creation system, we conducted a human evaluation study. A random sample of 20 characters was drawn from the total pool for assessment by five independent annotators. Prior to evaluation, annotators were instructed to use search engines and the provided script to familiarize themselves with each character's persona. They then rated the quality of the system-generated character profiles on a three-point Likert scale: 1 (Unsatisfactory), corresponding to a significant mismatch with the real persona; 2 (Acceptable), indicating general correctness despite minor inaccuracies; and 3 (Satisfactory), representing a highly accurate and well-summarized character profile.

The aggregated results showed that out of 100 total ratings (20 characters × 5 annotators), 55 were rated as ``Satisfactory," 38 as ``Acceptable," and only 7 as ``Unsatisfactory." This distribution strongly supports the efficacy of our character extraction pipeline.

\subsection{Evaluation Framework}

To comprehensively assess the model's role-playing capability, a dual-method evaluation framework is employed: metric-based evaluation and LLM-based evaluation. Metric-based evaluation quantifies the similarity between model-generated responses and corresponding ground-truth references. In contrast, LLM-based evaluation facilitates a deeper analysis of abstract qualities, such as role-playing fidelity and contextual coherence.

\subsubsection{Metric-based Evaluation}

Lexical and surface-form similarity between the generated text response and the ground truth is quantified using Rouge-L \cite{lin2004rouge} and Meteor \cite{banerjee2005meteor} scores, while semantic similarity is assessed via BertScore-F1 \cite{zhang2019bertscore}.  Additionally, the overall quality of the synthesized audio output is evaluated comprehensively using UTMOSv2 \cite{baba2024t05}.
\begin{table*}[htb]
  
  \label{tab:results_llm}
  \centering
   \begin{tabular}{lccccccc} %
    \toprule
    
    \textbf{Models} & \textbf{Likeness} & \textbf{Personality} & \textbf{Linguistic} & \textbf{Relation} & \textbf{Coherence} & \textbf{Acoustic} & \textbf{Overall} \\ %
    \midrule
    Qwen2.5-Omni \cite{xu2025qwen2}  & $3.54$ & $3.67$ & $3.97$ & $3.79$ & $4.33$ & $3.00$ & $3.72$ \\
    GLM-4-Voice \cite{zeng2024glm} & $3.46$ & $3.49$ & $3.70$ & $3.62$ & $3.98$ & $3.03$ & $3.55$ \\
    Baichuan-Audio \cite{li2025baichuan} & $2.96$ & $3.17$ & $3.47$ & $3.23$ & $3.68$ & $3.11$ & $3.27$ \\
    MiniCPM \cite{team2025minicpm} & $2.78$ & $2.94$ & $3.00$ & $3.01$ & $3.33$ & $2.90$ & $2.99$ \\
    Step-Audio \cite{huang2025step} & $3.33$ & $3.54$ & $3.74$ & $3.57$ & $3.98$ & $3.43$ & $3.60$ \\
    GPT-4o & $\mathbf{4.22}$ & $\mathbf{4.37}$ & $\mathbf{4.45}$ & $\mathbf{4.36}$ & $\mathbf{4.48}$ & $\mathbf{3.82}$ & $\mathbf{4.28}$ \\
    Gemini-2.5-flash & $3.62$ & $3.78$ & $4.01$ & $3.81$ & $4.16$ & $3.28$ & $3.78$ \\
    \bottomrule
  \end{tabular}
  \caption{LLM-based results on VoxRole}
\end{table*}
\subsubsection{LLM-based Evaluation}

Existing LLM-based paradigms predominantly rely on textual input, thereby neglecting the rich prosodic information inherent in spoken interactions. To address this limitation, we propose an acoustically-aware evaluation approach. First, we transcribe the model's spoken response into text using Whisper-3-Large. Then, we augment this textual representation with acoustic features extracted directly from the audio. Specifically, we utilize Emotion2Vec \cite{ma2023emotion2vec} to derive per-sentence emotion labels from both the dialogue context and the model's response audio. Furthermore, we extract the average pitch, energy, and speaking rate for each sentence and discretize these continuous values into categorical bins. The final input to the LLM judge\footnote{In this work, we use Gemini-2.5-flash as the LLM judge.} is a multimodal representation, concatenating the transcribed text with its corresponding acoustic features for both the context and the response.

The LLM judge then evaluates the model's performance along the following key dimensions:
\begin{enumerate}
    \item \textbf{Human-Likeness:} Does the response exhibit naturalness and expressiveness, conveying genuine personality, emotion, opinion, or doubt, or does it appear robotic and generic?
    \item \textbf{Personality Consistency:} How accurately does the response reflect the character's defined personality traits, background, and current mood?
    \item \textbf{Linguistic Fidelity:} Is the vocabulary, sentence structure, and tone consistent with the character's specified linguistic style?
    \item \textbf{Relational Coherence:} Does the response appropriately reflect the established relationship dynamics between the characters?
    \item \textbf{Contextual Coherence:} Is the response a logical and direct continuation of the immediately preceding dialogue turn, maintaining the conversation's flow?
    \item \textbf{Paralinguistic Appropriateness:} Are the specified vocal characteristics (e.g., pitch, energy, speaking rate, emotion label) congruent with the semantic content, emotional subtext of the scene, and the character's defined acoustic profile?
\end{enumerate}

The evaluation prompt incorporates a structured scoring rubric. High scores are awarded for responses demonstrating strong in-character fidelity and contextual relevance. Moderate scores are assigned to plausible but generic responses, while low scores are reserved for out-of-character, contradictory, or nonsensical replies.

\section{Experiments}

\subsection{Experimental Setup}

For our experiments, we evaluated a suite of state-of-the-art open-source speech dialogue models, including Baichuan-Audio \cite{li2025baichuan}, GLM-4-Voice \cite{zeng2024glm}, Step-Audio1 \cite{huang2025step}, Qwen2.5-Omni \cite{xu2025qwen2}, and MiniCPM-o2.6 \cite{team2025minicpm}. Additionally, we benchmarked the performance of the closed-source models GPT-4o and Gemini-2.5-flash on the role-playing task for reference.

The evaluation dataset was constructed from 20 movies randomly sampled from our corpus. From these movies, we extracted all dialogue segments consisting of six consecutive utterances. For each segment, the first five utterances served as the dialogue context (or history), while the final utterance was used as the prediction target (i.e., the ground truth).

Following the methodology of \cite{wang2025coser}, to provide the models with more comprehensive scene information, we employed a Large Language Model (LLM) to generate summaries of the scenario context. This summary was then combined with the role-playing instructions to form a unified system prompt, which was subsequently fed into the models under evaluation.

\subsection{Metric-based Results}

This metric-based evaluation delineates a distinct performance hierarchy among the assessed models, revealing critical insights into the capabilities of both proprietary and open-source systems. The analysis identifies GPT-4o as the preeminent model, demonstrating superior holistic performance across all evaluated dimensions. It achieved the highest scores on metrics for textual generation, including Rouge-L (12.91) and Meteor (18.30), indicating exceptional fluency and content relevance. Furthermore, its leading UTMOS score (3.66) confirms its state-of-the-art speech synthesis quality, characterized by high perceptual naturalness. While Gemini-2.5-flash exhibited competitive text generation capabilities, its performance in speech synthesis was markedly inferior to that of GPT-4o, creating a significant delta in multimodal output quality between the two leading proprietary models.

The open-source models presented more heterogeneous performance characteristics, illustrating clear trade-offs between semantic fidelity and speech quality. Notably, Step-Audio, the model with the largest parameter count (132B), achieved the highest BertScore F1 (84.16), signifying its strong proficiency in maintaining semantic accuracy. This strength, however, was coupled with a significant deficit in speech naturalness, as evidenced by its low UTMOS score (2.42). Conversely, the much smaller Qwen2.5-Omni (7B) demonstrated a remarkably well-balanced performance profile. Its most significant achievement was a speech naturalness score (UTMOS 3.57) that was highly competitive with, and nearly equivalent to, the top-performing GPT-4o. This finding suggests that advanced speech synthesis quality is not exclusively dependent on model scale. Other open-source models such as Baichuan-Audio and GLM-4-Voice yielded intermediate results, while MiniCPM consistently underperformed relative to its peers, particularly in the domain of speech synthesis.

A primary finding of this evaluation is the non-linear relationship between model parameter count and performance in key multimodal tasks. The observation that the 132B Step-Audio model was substantially outperformed in speech quality by the 7B Qwen2.5-Omni model challenges the prevailing assumption that performance scales monotonically with model size. This empirically substantiates the conclusion that factors such as model architecture, training methodologies, and data optimization play a critical role, and can in some cases be more influential than raw parameter scale in achieving high-quality, balanced multimodal output.

\subsection{LLM-based Result}

To provide a holistic assessment of model performance, we calculated the average score across the six dimensions, presented as the Overall score. The comprehensive evaluation across these six conversational dimensions reveals significant performance stratification, independent of model architecture. Results are shown in Table \ref{tab:results}

Coherence consistently emerges as the strongest capability across all systems. Every model scored above 3.33 in this dimension, with GPT-4o achieving near-human performance (4.48), indicating a relative maturity in maintaining dialogue flow and logical consistency. Conversely, Acoustic Quality represents the most pronounced universal limitation. Even the top-performing GPT-4o (3.82) is far from its peak performance in other areas, reflecting persistent, industry-wide challenges in audio generation that lag behind text understanding and generation capabilities.

A critical divergence appears in role-playing-specific dimensions, where a significant performance gap emerges between proprietary and open-source models. Personality Consistency and Relationship Understanding exhibit the largest of these gaps. GPT-4o (4.37/4.36) outperforms the best open-source model, Qwen2.5-Omni (3.67/3.79), by over 15\%, suggesting fundamental differences in their approaches to contextual modeling. In comparison, Linguistic Quality displays moderate variation, though smaller open-source models like MiniCPM (3.00) struggle significantly compared to their mid-sized counterparts. Notably, Qwen2.5-Omni demonstrates exceptional Coherence (4.33), surpassing even Gemini-2.5-flash (4.16), which challenges the conventional wisdom that model scale directly dictates performance in this dimension. Overall, the consistent underperformance of open-source models in the Personality and Relationship metrics highlights unresolved challenges in capturing and maintaining nuanced character dynamics.

\subsection{Subjective experimental results}

To validate the consistency between our LLM-based evaluation and human judgment, we conducted a human evaluation study on the VoxRole dataset. The primary objective of this study was to quantify the degree of agreement between our automated evaluation results and human expert judgments.

We randomly sampled 20 dialogue instances generated by five open-source models. Subsequently, 10 trained annotators were recruited to score the quality of these instances across six predefined metrics. They were provided with a detailed annotation manual that explicitly defined the six evaluation metrics, which mirrored those used in the LLM-based evaluation. We then calculated the Pearson correlation coefficient between the average human scores and the LLM-generated scores for these metrics. The analysis yielded a Pearson correlation coefficient of 0.762, indicating a strong positive correlation between the LLM's evaluations and human assessments. This result provides substantial evidence for the validity and reliability of our LLM-based evaluation method.

\begin{table}[htb]
  
  \centering
   \begin{tabular}{lcccc}
    \toprule
    
    {Context length} & \textbf{4} & \textbf{6} & \textbf{8} & \textbf{10} \\
    \midrule
    Qwen2.5-Omni \cite{xu2025qwen2}     &$3.69$ &$3.72$ & $3.70$ &$3.69$ \\
    GLM-4-Voice \cite{zeng2024glm}     &$3.54$ &$3.55$ & $3.57$& $3.56$\\
    Step-Audio \cite{huang2025step}      &$3.57$ &$3.60$ & $3.62$&$3.60$ \\
    \bottomrule
  \end{tabular}
  \caption{
  Ablation Study on Context Length
  }
  
  \label{tab:context_length}
\end{table}

\subsection{Ablation study on context length}

In our preliminary experiments, a fixed conversation length of 6 was utilized. To systematically investigate the influence of conversation length on the model's role-playing ability, we conducted a follow-up experiment by varying the context length. The average LLM score was employed as the evaluation metric. To ensure a fair comparison, the test set size and the prediction target were held constant across all experimental settings. The results, as presented in Table \ref{tab:context_length}, reveal a non-monotonic trend. Specifically, as the context length increases from 4 to 10, the LLM score initially improves and subsequently declines. This finding suggests that both insufficient and excessive contextual information can impair the model's overall performance, indicating the existence of an optimal context window.

\section{Conclusion}

To address the critical gap between text-based role-playing and the need for more immersive, speech-based conversational agents, this paper introduces VoxRole, the first comprehensive benchmark for speech-based Role-Playing Conversational Agents (RPCAs). Developed through a novel automated pipeline that extracts character-centric spoken dialogues from films and distills their multi-dimensional personas, VoxRole provides a robust resource for systematic evaluation. Our evaluation using this benchmark reveals significant performance disparities among current models, highlighting universal challenges in Paralinguistic Appropriateness and specific weaknesses in maintaining Personality Consistency and Relational Coherence, particularly in open-source systems. Future work will focus on expanding the dataset and fine-tuning large speech models to specifically enhance their role-playing capabilities.

\bibliography{aaai2026}

\end{document}